\documentclass[runningheads]{llncs}
\usepackage{graphicx}
\usepackage{amsmath,amssymb} 
\usepackage{color}
\usepackage[width=122mm,left=12mm,paperwidth=146mm,height=193mm,top=12mm,paperheight=217mm]{geometry}

\usepackage{mathrsfs}
\usepackage{algorithm}
\usepackage[noend]{algpseudocode}
\usepackage{color}
\usepackage{epstopdf}
\usepackage{subfig}
\usepackage{graphicx}
\usepackage{comment}

\DeclareMathOperator*{\argmin}{arg\,min}

\newcommand{\etal}{\textit{et al.}}
\newcommand{\ie}{\textit{i.e.}}

\begin{document}
\pagestyle{headings}
\mainmatter
\def\ECCV18SubNumber{1648}  

\title{Fast Convolutional Sparse Coding in the Dual Domain} 

\titlerunning{Fast Convolutional Sparse Coding in the Dual Domain}

\authorrunning{L. Affara, B. Ghanem, P. Wonka}

\author{Lama Affara, Bernard Ghanem, Peter Wonka}
\institute{King Abdullah University of Science and Technology (KAUST), Saudi Arabia\\
        \email{lama.affara@kaust.edu.sa, bernard.ghanem@kaust.edu.sa,
        pwonka@gmail.com}
}

\maketitle

\begin{abstract}
   Convolutional sparse coding (CSC) is an important building block of many computer vision applications ranging from image and video compression to deep learning.
We present two contributions to the state of the art in CSC. First, we significantly speed up the computation by proposing a new optimization framework that tackles the problem in the dual domain. Second, we extend the original formulation to higher dimensions in order to process a wider range of inputs, such as RGB images and videos. Our results show up to 20 times speedup compared to current state-of-the-art CSC solvers.
\end{abstract}

\section{Introduction}
\label{sec:intro}

Human vision is characterized by the response of neurons to stimuli within their receptive fields, which is usually modeled mathematically by the convolution operator. Correspondingly for computer vision, coding the image based on a convolutional model has shown its benefits through the development and application of deep Convolutional Neural Networks. Such a model constitutes a strategy for unsupervised feature learning, and more specifically to patch-based feature learning also known as dictionary learning.\\
\indent Convolutional Sparse Coding (CSC) is a special type of sparse dictionary learning algorithms. It uses the convolution operator in its image representation model rather than generic linear combinations. This results in diverse translation-invariant patches and maintains the latent structures of the underlying signal. CSC has recently been applied in a wide range of computer vision problems such as image and video processing~\cite{elad2006image,aharon2006uniqueness,couzinie2011dictionary,yang2008image,Gu2016}, structure from motion~\cite{zhu2015convolutional}, computational imaging~\cite{heide2014imaging}, tracking ~\cite{Zhang2016}, as well as the design of deep learning architectures~\cite{krizhevsky2012imagenet}.\\
\indent Finding an efficient solution to the CSC problem however is a challenging task due to its high computational complexity and the non-convexity of its objective function. Seminal advances~\cite{Bristow2013,Kong2014,Heide2015} in CSC have shown computational speed-up by solving the problem efficiently in the Fourier domain where the convolution operator is transformed to element-wise multiplication. As such, the optimization is modeled as a biconvex problem formed by two convex subproblems, the coding subproblem and the learning subproblem, that are iteratively solved in a fixed point manner.\\
\indent Despite the performance boost attained by solving the CSC optimization problem in the Fourier domain, the problem is still computationally expensive due to the dominating cost of solving large linear systems. More recent work ~\cite{Heide2015,Kong2014} makes use of the block-diagonal structure of the matrices involved and solves the linear systems in a parallel fashion, thus leveraging hardware acceleration.\\
\indent Inspired by recent work on circulant sparse trackers~\cite{Zhang2016}, we model the CSC problem in the dual domain. The dual formulation casts the coding subproblem into an  Alternating Direction Method of Multipliers (ADMM) framework that involves solving a linear system with a lower number of parameters than previous work. This allows our algorithm to achieve not only faster convergence towards a feasible solution, but also a lower computational cost. The solution for the learning subproblem in the dual domain is achieved by applying coordinate ascent over the Lagrange multipliers and the dual parameters. Our extensive experiments show that the dual framework achieves significant speedup over the state of the art while converging to comparable objective values.\\
\indent Moreover, recent work on higher order tensor formulations for CSC (TCSC) handles the problem with an arbitrary order tensor of data which allows learning more elaborate dictionaries such as colored dictionaries. This allows a richer image representation and greatly benefits the applicability of CSC in other application domains such as color video reconstruction. Our dual formulation provides faster performance compared to TCSC by eliminating the need for solving a large number of linear systems involved in the coding subproblem which dominates the cost for solving the problem.\\

\noindent\textbf{Contributions.}~ We present two main contributions. (1) We formulate the CSC problem in the dual domain and show that this formulation leads to faster convergence and thus lower computation time. (2) We extend our dual formulation to higher dimensions and gain up to 20 times speedup compared to TCSC.

\section{Related Work}
\label{sec:related}
\begin{figure*}[t]
	\centering
		\includegraphics[width=0.75\linewidth]{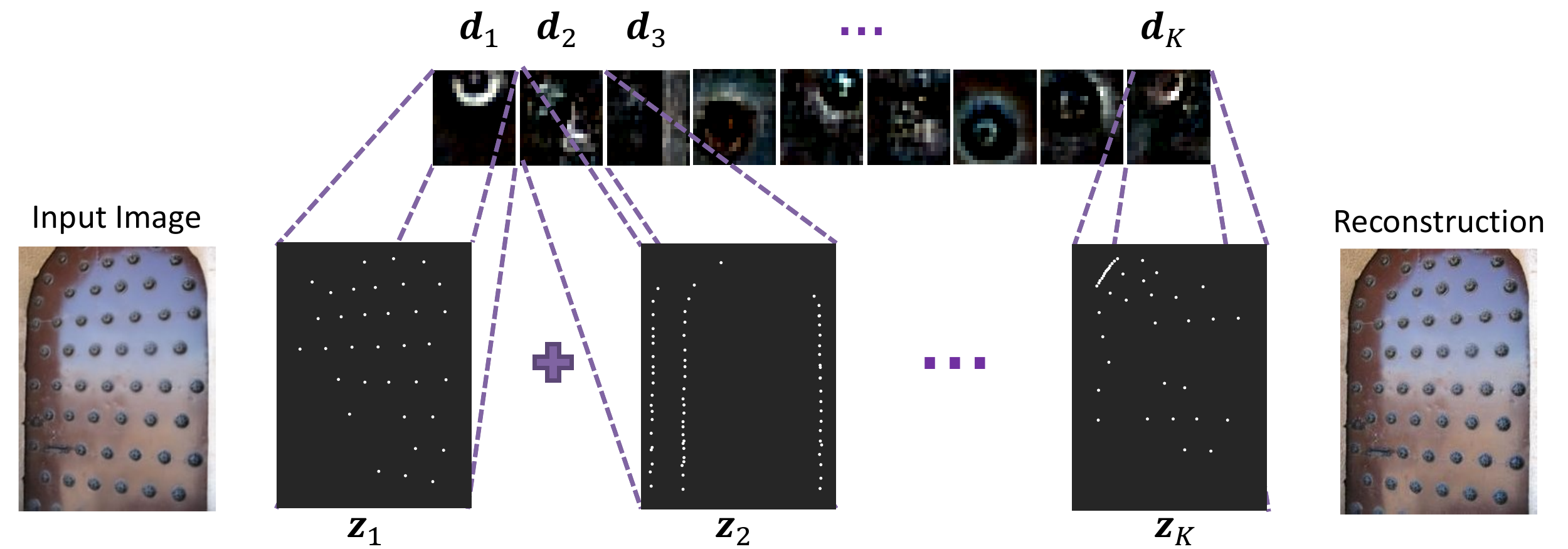}
	\centering
	\caption{Convolutional Sparse Coding Model. An input image is reconstructed as a sum of dictionary elements convolved with their corresponding sparse maps.}
	\label{fig:CSC_main}
\end{figure*}
As mentioned earlier, CSC has many applications and quite a few methods have been proposed to solve the non-convex CSC optimization. In the following, we mainly review the works that focus on the computational complexity and efficiency aspects of the problem.

The seminal work of~\cite{Zeiler2010} proposes \emph{Deconvolutional Networks}, a learning framework based on convolutional decomposition of images under a sparsity constraint. Unlike previous work in sparse image decomposition~\cite{olshausen1997sparse,lee2006efficient,mairal2009online,mairal2009supervised} that builds hierarchical representations of an image on a patch level, \emph{Deconvolutional Networks} perform a sparse decomposition over entire images. This strategy significantly reduces the redundancy among filters compared with those obtained by the patch-based approaches. Kavukcuoglu \etal~\cite{kavukcuoglu2010learning}
propose a convolutional extension to the coordinate descent sparse coding algorithm~\cite{li2009coordinate} to represent images using convolutional dictionaries for object recognition tasks. Following this path, Yang \etal~\cite{yang2010supervised} propose a supervised dictionary learning approach to improve the efficiency of sparse coding.

To efficiently solve the complex optimization problems in CSC, most  existing solvers attempt to transform the problem into the frequency domain.
\v{S}orel and \v{S}roubek~\cite{vsorel2016fast} propose a non-iterative method for computing the inversion of the convolutional operator in the Fourier domain using the matrix inversion lemma. Bristow \etal~\cite{Bristow2013} propose a quad-decomposition of the original objective into convex subproblems and  exploit the ADMM approach to solve the convolution subproblems in the Fourier domain. In their follow-up work~\cite{Bristow2014}, a number of optimization methods for solving convolution problems and their applications are discussed. In the work of~\cite{Kong2014}, the authors further exploit the separability of convolution across bands in the frequency domain. Their gain in efficiency arises from  computing a partial vector (instead of a full vector). To further improve efficiency, Heide \etal~\cite{Heide2015} transform the original constrained problem into an unconstrained problem by encoding the constraints in the objective using some indicator functions. The new objective function is then further split into a set of convex functions that are easier to optimize separately. They also devise a more flexible solution by adding a diagonal matrix to the objective function to handle the boundary artifacts resulting from transforming the problem into the Fourier domain.

Various CSC methods have also been proposed for different applications. Zhang \etal~\cite{Zhang2016} propose an efficient sparse coding method for sparse tracking. They also solve the problem in the Fourier domain, in which the $\ell_1$ optimization is obtained by solving its dual problem and thus achieving more efficient computation. Unlike traditional sparse coding based image super resolution methods that divide the input image into overlapping patches, Gu \etal~\cite{Gu2016} propose to decompose the image by filtering. Their method is capable of reconstructing local image structures. Similar to~\cite{Bristow2013}, the authors also solve the subproblems in the Fourier domain. The stochastic average and ADMM algorithms~\cite{zhong2014fast} are used for a memory efficient solution. Recent work \cite{Wang2017OnlineCS,wohlberg2016boundary,wohlberg2016convolutional} has also reformulated the CSC problem by extending its applicability to higher dimensions ~\cite{bibi2017high} and to large scale data~\cite{choudhury2017consensus}.

In this work, we attempt to provide a more efficient solution to CSC and higher order CSC by tackling the optimization problem in its dual form.


\section{CSC Formulation and Optimization}
In this section, we present the mathematical formulation of the CSC problem and show our approach to solving its subproblems in their dual form. There are multiple slightly different, but similar formulations for the CSC problem. Heide \etal~\cite{Heide2015} introduced a special case for boundary handling, but we use the more general formulation that is used by most authors. Thus, unlike~\cite{Heide2015}, we assume circular boundary conditions in our derivation of the problem. Brisow \etal \cite{Bristow2013} verified that this assumption has a negligible effect for small support filters, which is generally the case in dictionary learning where the learned patches are of a small size relative to the size of the image. In addition, they show that the Fourier transform can be replaced by the Discrete Cosine Transform when the boundary effects are problematic.


\subsection{CSC Model}
The CSC problem is generally expressed in the form
\begin{align}
\label{eq:CSC}
\begin{split}
\argmin_{\mathbf{d},\mathbf{z}}&\;\frac{1}{2}\|\mathbf{x}-\sum_{k=1}^K \mathbf{d}_k * \mathbf{z}_k \|_2^2 + \beta \sum_{k=1}^K \|\mathbf{z}_k\|_1\\
\text{subject to}&\;\; \|\mathbf{d}_k\|_2^2 \leq 1 \;\;\forall k \in \{1,...,K\}
\end{split}
\end{align}
where $\mathbf{d}_k \in \mathbb{R}^M$ are the vectorized 2D patches representing $K$ dictionary elements, and $\mathbf{z}_k \in \mathbb{R}^D$ are the vectorized sparse maps corresponding to each of the dictionary elements (see Figure~\ref{fig:CSC_main}). The data term represents the image $\mathbf{x} \in \mathbb{R}^D$ modelled by the sum of convolutions of the dictionary elements with their corresponding  sparse maps, and $\beta$ controls the tradeoff between the sparsity of the feature maps and reconstruction error. The inequality constraint on the dictionary elements assumes Laplacian distributed coefficients, which ensures solving the problem at a proper scale for all elements since a larger value of $\mathbf{d}_k$ would scale down the value of the corresponding $\mathbf{z}_k$. The above equation shows the objective function for a single image, and it can be easily extended to multiple images, where $K$ corresponding sparse maps are inferred for each image and all the images share the same $K$ dictionary elements.

\subsubsection{CSC Subproblems}
The objective in Eq.~\ref{eq:CSC} is not jointly convex. However, using a fixed point approach (\ie iteratively solving for one variable while keeping the other fixed) leads to two convex subproblems, which we refer to as the coding subproblem and the dictionary learning subproblem. For ease of notation, we represent the convolution operations by multiplication of Toeplitz matrices with the corresponding variables.

\noindent\textbf{Coding Subproblem.} We infer the sparse maps for a fixed set of dictionary elements as shown in Eq.~\ref{eq:Inference}.
\begin{equation}
\label{eq:Inference}
\argmin_{\mathbf{z}}\;\;\frac{1}{2} \|\mathbf{x}-\mathbf{D}\mathbf{z}\|_2^2 + \beta\|\mathbf{z}\|_1\\
\end{equation}
Here, $\mathbf{D}=[\mathbf{D}_1\dots\mathbf{D}_K]$ is of size $D \times DK$ and is a concatenation of the convolution matrices of the dictionary elements, and $\mathbf{z}=[\mathbf{z}_1^T\dots \mathbf{z}_K^T]^T$ is a concatenation of the vectorized sparse maps.\\

\noindent\textbf{Learning Subproblem.} We learn the dictionary elements for a fixed set of sparse feature maps as shown in Eq.~\ref{eq:Learning}.
\begin{align}
\label{eq:Learning}
\begin{split}
\argmin_{\mathbf{d}}&\;\;\frac{1}{2}\|\mathbf{x}-\mathbf{Z}\mathbf{S}^T\mathbf{d}\|_2^2\\
\text{subject to}&\;\; \|\mathbf{d}_k\|_2^2 \leq 1 \;\;\forall k \in \{1,...,K\}
\end{split}
\end{align}
Similar to above, $\mathbf{Z}=[\mathbf{Z}_1\dots\mathbf{Z}_K]$ is of size $D \times DK$ and is a concatenation of the sparse convolution matrices, $\mathbf{d}=[\mathbf{d}_1^T\dots \mathbf{d}_K^T]^T$ is a concatenation of the dictionary elements, and $\mathbf{S}$ projects the filter onto its spatial support. 

The above two subproblems can be optimized iteratively using ADMM \cite{Bristow2013,Heide2015}, where  each ADMM iteration requires solving a large linear system of size $DK$ for each of the two variables $\mathbf{z}$ and $\mathbf{d}$. Moreover, when applied to multiple images, solving the linear systems for the coding subproblem can be done separably, but should be done jointly for the learning subproblem, since all images share the same dictionary elements (see Section \ref{sec:complexity_analysis} for more details on complexity analysis).

\subsection{CSC Dual Optimization}
In this section, we show our approach to solving the CSC subproblems in the dual domain. Formulating the problems in the dual domain reduces the number of parameters involved in the linear systems from $DK$ to $D$, which leads to faster convergence towards a feasible solution and thus better computational performance. Since the two subproblems are convex, the duality gap is zero and solving the dual problem is equivalent to solving the primal form. In addition, similar to \cite{Bristow2013}, we also solve the convolutions efficiently in the Fourier domain as described below.

\subsubsection{Coding Subproblem}
To find the dual problem of Eq.~\ref{eq:Inference}, we first introduce a dummy variable $\mathbf{r}$ with equality constraints to yield the following formulation
\begin{align}
\begin{split}
\min_{\mathbf{z},\mathbf{r}}&\;\;\frac{1}{2}\|\mathbf{r}\|_2^2 + \beta\|\mathbf{z}\|_1\\
\text{subject to}&\;\;\mathbf{r}=\mathbf{D}\mathbf{z}-\mathbf{x}
\end{split}
\end{align}
The Lagrangian of this problem would be:
\begin{equation}
\mathcal{L} \left ( \mathbf{z},\mathbf{r},\boldsymbol{\lambda} \right )=\frac{1}{2}\|\mathbf{r}\|_2^2 + \beta\|\mathbf{z}\|_1+\boldsymbol{\lambda}^T\left(\mathbf{D}\mathbf{z}-\mathbf{x}-\mathbf{r}\right)
\end{equation}
which results in the following dual optimization with  dual variable $\boldsymbol{\lambda}$:
\begin{equation}
\max_{\boldsymbol{\lambda}}\;\;\left[
\min_{\mathbf{r}} \frac{1}{2}\|\mathbf{r}\|_2^2 -\boldsymbol{\lambda}^T\mathbf{x} -\boldsymbol{\lambda}^T\mathbf{r} +
\min_{\mathbf{z}} \beta\|\mathbf{z}\|_1+\boldsymbol{\lambda}^T\mathbf{D}\mathbf{z}\right]
\end{equation}
Solving the minimizations over $\mathbf{r}$ and $\mathbf{z}$ and using the definition of the conjugate function to the $l_1$ norm, we get the dual problem of Eq.~\ref{eq:Inference} as:
\begin{align}
\begin{split}
\label{eq:Inference_Dual}
\min_{\boldsymbol{\lambda}}&\;\;\frac{1}{2}\boldsymbol{\lambda}^T\boldsymbol{\lambda}+\boldsymbol{\lambda}^T\mathbf{x}\\
\text{subject to}&\;\;\|\mathbf{D}^T\boldsymbol{\lambda}\|_\infty \leq \beta
\end{split}
\end{align}

\noindent\textbf{Coding Dual Optimization.} Now, we show how to solve the optimization problem in Eq.~\ref{eq:Inference_Dual} using ADMM. ADMM generally solves convex optimization problems by breaking the original problem into easier subproblems that are solved iteratively. To apply ADMM here, we introduce an additional variable $\boldsymbol{\theta}=\mathbf{D}^T\boldsymbol{\lambda}$,  which allows us to write the problem in the general ADMM form as shown in Eq.~\ref{eq:Inference_ADMM}. Since the dual solution to a dual problem is the primal solution for convex problems, the Lagrange multiplier involved in the ADMM update step is the sparse map vector $\mathbf{z}$ in Eq.~\ref{eq:Inference}.
\begin{align}
\begin{split}
\label{eq:Inference_ADMM}
\min_{\boldsymbol{\lambda},\boldsymbol{\theta}}&\;\;h\left(\boldsymbol{\lambda}\right) + g\left(\boldsymbol{\theta}\right) \text{ s.t. } \boldsymbol{\theta}=\mathbf{D}^T\boldsymbol{\lambda}\\
\text{where } h\left(\boldsymbol{\lambda}\right)&=\frac{1}{2}\boldsymbol{\lambda}^T\boldsymbol{\lambda}+\boldsymbol{\lambda}^T\mathbf{x} \text{ and }  g(\boldsymbol{\theta})=ind_C\left(\boldsymbol{\theta}\right)
\end{split}
\end{align}
Here, $ind_C(.)$ is the indicator function defined on the convex set of constraints $C=\{\boldsymbol{\theta}\;|\;\|\boldsymbol{\theta}\|_\infty \leq \beta\}$. Deriving the augmented Lagrangian of the problem and solving for the ADMM update steps~\cite{boyd2011distributed} yields the following iterative solutions to the dual problem with $i$ representing the iteration number.
\begin{align}
\begin{split}
\label{eq:Inference_solution}
\boldsymbol{\lambda}^{i+1}&=( \mathbf{D}\mathbf{D}^T + \frac{1}{\rho}\mathbf{I} )^{-1} ( \mathbf{D}\boldsymbol{\theta}^i + \frac{1}{\rho}\mathbf{D}\mathbf{z}^i -\frac{1}{\rho}\mathbf{x} )\\
\boldsymbol{\theta}^{i+1}&=\Pi_C ( \mathbf{D}^T\boldsymbol{\lambda}^{i+1}-\frac{1}{\rho}\mathbf{z}^i  )\\
\mathbf{z}^{i+1}&= \mathbf{z}^i + \rho(\boldsymbol{\theta}^{i+1}-\mathbf{D}^T\boldsymbol{\lambda}^{i+1})
\end{split}
\end{align}
The parameter $\rho\in\mathbb{R}+$  denotes the step size for the ADMM iterations, and $\Pi$ represents the projection operator onto the set $C$. The linear systems shown above do not require expensive matrix inversion or multiplication as they are transformed to elementwise divisions and multiplications when solved in the Fourier domain. This is possible because ignoring the boundary effects leads to a circulant structure for the convolution matrices, and thus they can be expressed by their base sample as follows:
\begin{equation}
\mathbf{D_k}=\mathcal{F}diag(\hat{\mathbf{d_k}})\mathcal{F}^H
\end{equation}
where $\hat{\mathbf{d}}$ denotes the Discrete Fourier Transform (DFT) of $\mathbf{d}$, $\mathcal{F}$ is the DFT matrix independent of $\mathbf{d}$, and $X^H$ is the Hermitian transpose.

\begin{figure}[t]
	\centering
        \includegraphics[width=1\linewidth]{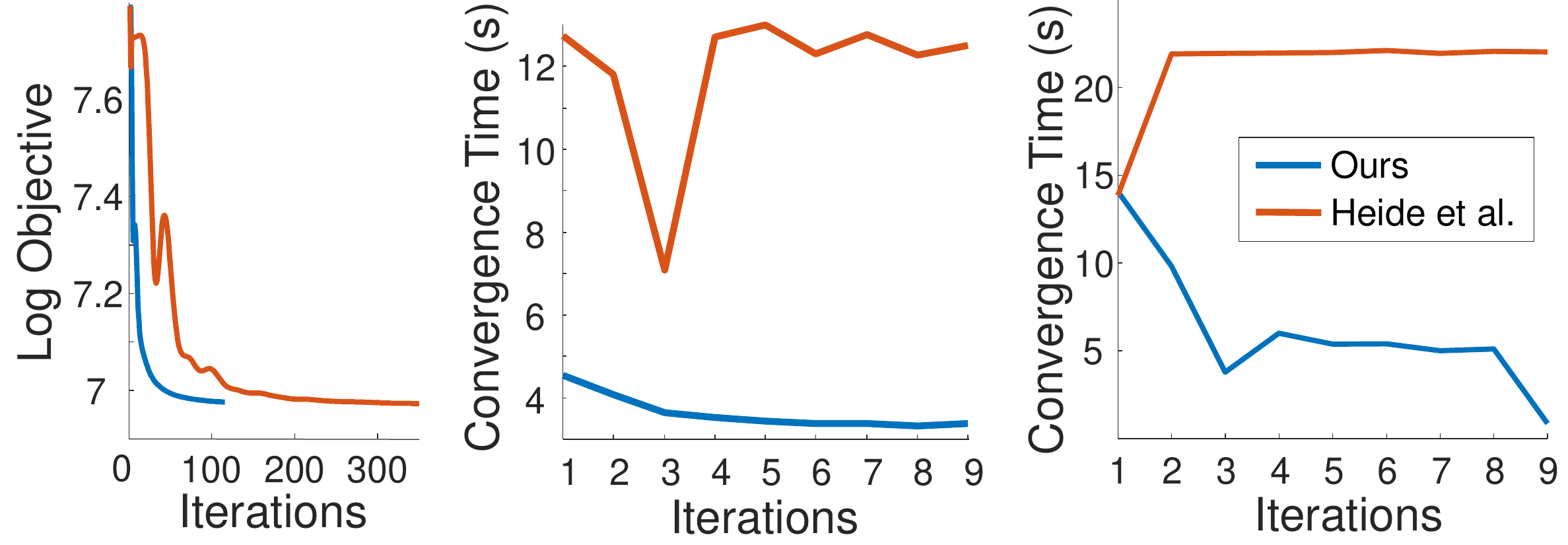}
	\caption{Convergence (left) and computation time (middle) of the coding subproblem. Computation time of the learning subproblem (right).}
	\label{fig:learning_coding_conv}
\end{figure}

In our formulation, the $\boldsymbol{\lambda}$-update step requires solving a linear system of size $D$. Heide \etal~\cite{Heide2015} however solve the problem in the primal domain in which the $\mathbf{d}$-update step involves solving a much larger system of size $KD$. Clearly, our solution in the dual domain will  lead to faster ADMM convergence. Figure \ref{fig:learning_coding_conv}-left shows the coding subproblem convergence of our approach and that of Heide \etal Our dual formulation leads to convergence within less iterations compared to the primal domain. In addition, our approach achieves a lower objective in general at any feasible number of iterations.

\subsubsection{Learning Subproblem}
To find a solution for Eq.~\ref{eq:Learning}, we minimize the Lagrangian of the problem (see Eq.~\ref{eq:Learning_Lagrangian}) assuming optimal values for the Lagrange multipliers $\mu_k$. This results in the optimization problem shown in Eq.~\ref{eq:Learning_new}.
\begin{equation}
\label{eq:Learning_Lagrangian}
\mathcal{L}(\mathbf{d},\boldsymbol{\mu})=\frac{1}{2}\|\mathbf{x}-\mathbf{Z}\mathbf{S}^T\mathbf{d}\|_2^2+\sum_{k=1}^K\mu_k(\|\mathbf{d}_k\|_2^2-1)
\end{equation}

\begin{equation}
\label{eq:Learning_new}
\argmin_{\mathbf{d}}\frac{1}{2}\|\mathbf{x}-\mathbf{Z}\mathbf{S}^T\mathbf{d}\|_2^2+\sum_{k=1}^K\mu_k^*\|\mathbf{d}_k\|_2^2
\end{equation}
To find the dual problem of Eq.~\ref{eq:Learning_new}, we follow a similar approach to the inference subproblem by introducing a dummy variable $\mathbf{r}$ with equality constraints such that $\mathbf{r}=\mathbf{Z}\mathbf{S}^T\mathbf{d}-\mathbf{x}$. Deriving the Lagrangian of the problem and minimizing over the primal variables yields the dual problem shown in~\ref{eq:Learning_Dual}.
\begin{equation}
\label{eq:Learning_Dual}
\min_{\boldsymbol{\gamma}}\;\;\frac{1}{2}\boldsymbol{\gamma}^T\boldsymbol{\gamma}+\boldsymbol{\gamma}^T\mathbf{x}+\sum_{k=1}^K\frac{1}{4\mu_k^*}\|\mathbf{S}_k\mathbf{Z}_k^T\boldsymbol{\gamma}\|_2^2
\end{equation}

\noindent\textbf{Learning Dual Optimization.}
The optimization problem in Eq.~\ref{eq:Learning_Dual} has a closed form solutionas shown in Eq.~\ref{eq:Learning_gamma}.
\begin{equation}
\label{eq:Learning_gamma}
\boldsymbol{\gamma}^*=-\left(\mathbf{I}+\sum_{k=1}^K\frac{1}{2\mu_k^*}\mathbf{Z}_k\mathbf{S}^T_k\mathbf{S}_k\mathbf{Z}_k^T\right)^{-1}\mathbf{x}
\end{equation}
Given the optimal value for the dual variable $\boldsymbol{\gamma}$, we can compute the optimal value for the primal variable $\mathbf{d}$ as follows:
\begin{equation}
\label{eq:Learning_Solution}
\mathbf{d}_k^*=-\frac{1}{2\mu_k^*}\mathbf{S}_k\mathbf{Z}_k^T\boldsymbol{\gamma}^*\;\;\forall k=\{1,...,K\}
\end{equation}

To find the optimal values for the Lagrange multipliers $\mu_k$, we need to assure that the KKT conditions are satisfied. At optimal $(\mu_k^*,\mathbf{d}_k^*)$, the solution to the primal problem in Eq.~\ref{eq:Learning} and its Lagrangian are equal. Thus, we end up with the below iterative update step for $\mu_k$. 
\begin{equation}
\label{eq:mu_update}
\mu_k^{i+1}=\mu_k^i \|\mathbf{d}_k^i\|_2
\end{equation}
The learning subproblem is then solved iteratively by alternating between updating $\mathbf{d_k}$ as per Eqs.~\ref{eq:Learning_gamma},\ref{eq:Learning_Solution} and updating $\mu_k$ as per Eq.~\ref{eq:mu_update} until convergence is achieved.

We use conjugate gradient to solve the system involved in the $\boldsymbol{\gamma}$-update step by applying the heavy convolution matrix multiplications in the Fourier domain. The computation cost for solving this system decreases with ADMM iterations, since we employ a warm start where we  initialize  $\boldsymbol{\gamma}$ with the solution  from the previous iteration. Figure~\ref{fig:learning_coding_conv}-right shows the decreasing computation time of the learning subproblem of our approach.

For more details on the derivations of the coding and learning subproblems as well as the solutions to the equations in the Fourier domain, you may refer to the \textbf{supplementary material}.
\subsubsection{Coordinate Descent}
Now that we derived a solution to the two subproblems, we can use coordinate descent to solve the joint objective in Eq.~\ref{eq:CSC} by alternating between the solutions for $\mathbf{z}$ and $\mathbf{d}$. The full algorithm for the CSC problem is shown in Alg.~\ref{alg:CSC}.\\
\begin{algorithm}
\begin{algorithmic}[1]
\State Set ADMM optimization parameter $\rho>0$
\State Initialize variables $\mathbf{z}$, $\mathbf{d}$, $\boldsymbol{\theta}$,  $\boldsymbol{\mu}$
\State Apply FFT $\rightarrow\mathbf{\hat{z}}$, $\mathbf{\hat{d}}$, $\boldsymbol{\hat{\theta}}$, $\mathbf{\hat{x}}$
\While {not converged}
\State \parbox[t]{\dimexpr\linewidth-\algorithmicindent}{Update $\mathbf{\hat{z}}$, $\mathbf{z}$, $\boldsymbol{\hat{\theta}}$, $\boldsymbol{\theta}$, $\boldsymbol{\hat{\lambda}}$, $\boldsymbol{\lambda}$ iteratively by solving Eq.~\ref{eq:Inference_solution} in the Fourier domain when possible\strut}
\State \parbox[t]{\dimexpr\linewidth-\algorithmicindent}{Update $\boldsymbol{\gamma}, \mathbf{d}$ using Eq.~\ref{eq:Learning_gamma},~\ref{eq:Learning_Solution}\strut}
\State \parbox[t]{\dimexpr\linewidth-\algorithmicindent}{Update $\mu_k^{i+1}=\mu_k^i\|d_k^i\|_2\;\forall k \in \{1,...,K\}$\strut}
\EndWhile
\State Output solution variables by inverse FFT
\end{algorithmic}
\caption{Convolutional Sparse Coding}
\label{alg:CSC}
\end{algorithm}

The coordinate descent algorithm above guarantees a monotonically decreasing joint objective. We keep iterating until convergence is reached, \ie when the change in the objective value, or the solution for the optimization variables $\mathbf{d}$ and $\mathbf{z}$ reaches a user-defined threshold $\tau=10^{-3}$. For solving the coding and learning subproblems, we also run the algorithms until convergence.

\subsection{Higher Order Tensor CSC}
Higher order tensor CSC~\cite{bibi2017high} allows convolutional sparse coding over higher dimensional input such as a set of 3D input images as well as 4D videos. Similar to before, given the input data, we seek to reconstruct using $K$ patches convolved with $K$ sparse maps. In this formulation, each of the patches is of the same order of dimensionality as the input data with the possibility of a smaller spatial support. In addition, TCSC allows high dimensional correlation among features in the data. In this sense, unlike traditional CSC in which a separate sparse code is learned for separate features, the sparse maps in TCSC are shared along one of the dimensions such as the color channels for images/videos. The reader may be referred to the paper by Bibi \etal~\cite{bibi2017high} for more details of the derivations for TCSC. Below we give our approach to solving the TCSC coding and learning subproblems in the dual domain. \\

The dictionary elements are represented by a tensor $\mathcal{D}$ $\in R^{J\times K\times n_1 \times ... \times n_d}$ where $J$ represents the correlated input dimension (usually referring to data features/channels), $K$ is the number of elements, and $n_1,...,n_d$ are the uncorrelated dimensions (\ie  $\;d=2$ representing the spatial dimensions for images, and $d=3$ representing the spatial and time dimensions for videos). In our dual formulation, we perform circulant tensor unfolding~\cite{bibi2017high} resulting in a block circulant dictionary matrix $\mathbf{D}$ of size $JD\times KD$ where $D= n_1 \times ... \times n_d$.  Thus, each of the $K$ convolution matrices is now of size $DJ\times D$ where $\mathbf{D}_k=[\mathbf{D}_{1,k}^T\dots\mathbf{D}_{J,k}^T]^T$. \\

The solution to the coding dual problem is shown in Eq.~\ref{eq:Inference_solution} where the inverse in the $\lambda$-update step involves now a block diagonal matrix. Thus, the inversion can be done efficiently by parallelization over the $D$ blocks while making use of the Woodburry inversion formula~\cite{Bristow2013,Heide2015}.

The solution to the dictionary learning subproblem is straightforward, since the solution is separable along the dimension $J$. We solve for $\boldsymbol{\gamma}_j^*$ and $\mathbf{d}_{k,j}^*$ as previously shown in Eqs.~\ref{eq:Learning_gamma} and~\ref{eq:Learning_Solution} for all $\mathbf{x}_j$ where $j=1\dots J$.

\section{Results}
\label{results}
In this section, we give an overview of the implementation details and the parameters selected for our dual CSC solver. We also show the complexity analysis and convergence of our approach compared to~\cite{Heide2015}, the current state-of-the-art CSC solver. Finally, we show results on 4D TCSC using color input images as well as 5D TCSC using colored videos.

\subsection{Implementation Details}
\label{sec:implementation}
We implemented the algorithm in MATLAB using the Parallel Computing Toolbox and we ran the experiments on an Intel 3.1GHz processor machine. We used the code provided by~\cite{Heide2015} in the comparisons for regular CSC and by~\cite{bibi2017high} in the comparisons to TCSC. We evaluate our approach on the \emph{fruit} and \emph{city} datasets formed of 10 images each, the \emph{house} dataset containing 100 images, and \emph{basketball} video from the OTB50 dataset selecting 10 frames similar to~\cite{bibi2017high}.

We apply contrast normalization to the images prior to learning the dictionaries for both gray scale and color images; thus, the figures show normalized patches. We show results by varying the sparsity coefficient $\beta$, the number of dictionary elements $K$, and  the number of images $N$. In our optimization, we choose a constant value of $\rho=0.1$ for the ADMM step size. We also initialize $\mathbf{z}$ and $\boldsymbol{\theta}$ with zeros for the first iteration of the learning subproblem, and $\mathbf{d}$ with random values in the coding subproblem. Our results compare with Heide \etal~\cite{Heide2015} for regular CSC, as it is the fastest among the published methods discussed in the related work section, and with Bibi \etal~\cite{bibi2017high} for TCSC as it is the only method that deals with higher order CSC.

\begin{figure*}[t]
	\centering
		\includegraphics[width=\textwidth]{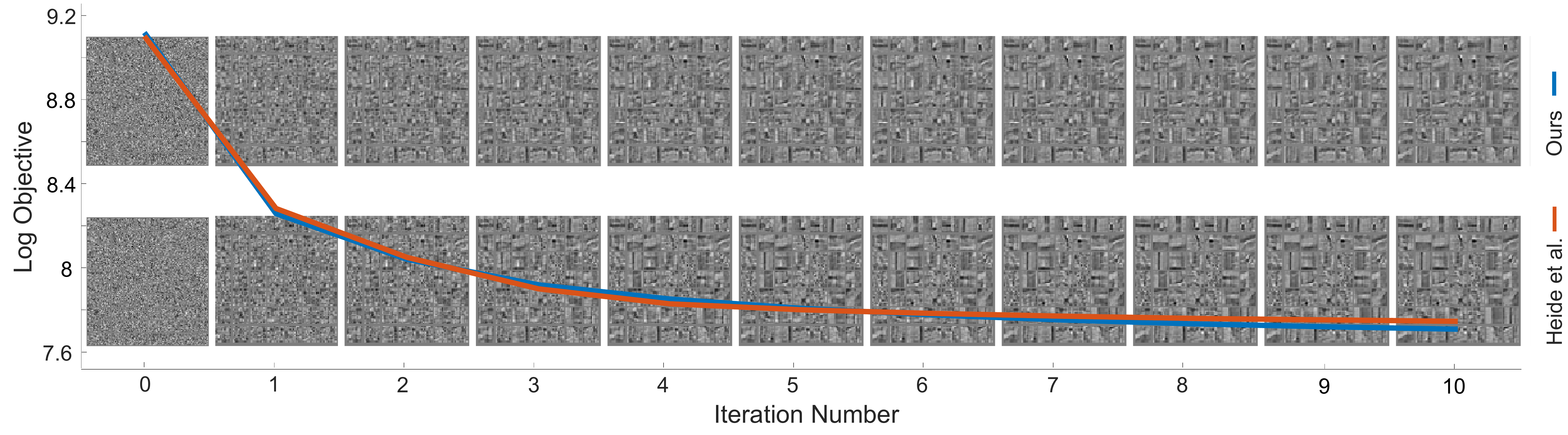}
	\centering
	\caption{Dictionary learning convergence versus number of iterations, comparing our method (top) and Heide \etal~\cite{Heide2015} (bottom). Starting from the same initial dictionary shown at iteration $0$, the objective value decreases as the number of iterations increase. Each point in the graph is in correspondence with a set of dictionary elements.}
	\label{fig:Obj_Iter}
\end{figure*}

\subsection{Complexity Analysis}
\label{sec:complexity_analysis}

In this section, we analyze the per-iteration complexity of our approach compared to~\cite{Heide2015} and~\cite{bibi2017high} with respect to each of the subproblems as shown in Table~\ref{complexitytable}. In the equations below, $D$ corresponds to the product of the order of the uncorrelated dimensions (e.g. number of pixels for images), $J$ is the number of channels in the correlated  input  dimension, and $Q$ is the number of conjugate gradient iterations within the learning subproblem. For regular CSC on high dimensional data, we assume that the problem is solved separately for each of the $J$ channels.

\noindent\textbf{Coding Subproblem.} In the coding subproblem, the complexity of our approach is similar to that of~\cite{Heide2015} for regular 2D CSC ($J=1$). The computational complexity is dominated by solving the linear system by elementwise product and division operations in the Fourier domain. Although  the two approaches are computationally similar here, it is important to note  that the number of variables involved in solving the systems is much less in our approach. In practice, we observe that this also leads to faster convergence for the subproblem as shown in Figure~\ref{fig:learning_coding_conv}. For higher order dimensional CSC ($J>1$), our formulation is linear in the number of filters compared to a cubic cost for TCSC. In TCSC, the Sherman Morrison formula no longer applies and the computation is dominated by solving $D$ linear systems of size $K\times K$.

\begin{figure}[t]
	\centering
	\subfloat[][]{
        \includegraphics[width=0.315\linewidth]{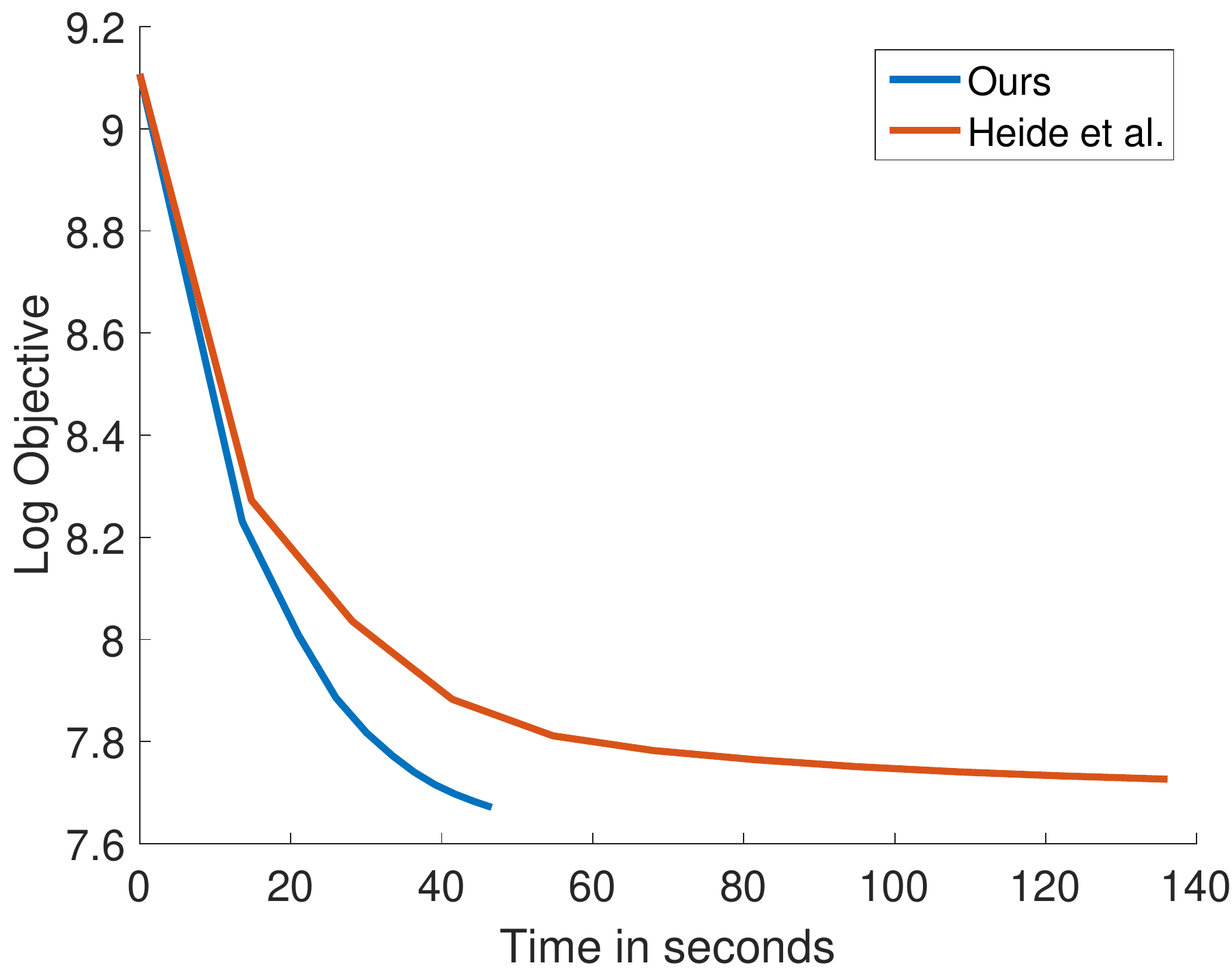}
        \label{fig:Obj_Time}
    }
	\subfloat[][]{
        \includegraphics[width=0.315\linewidth]{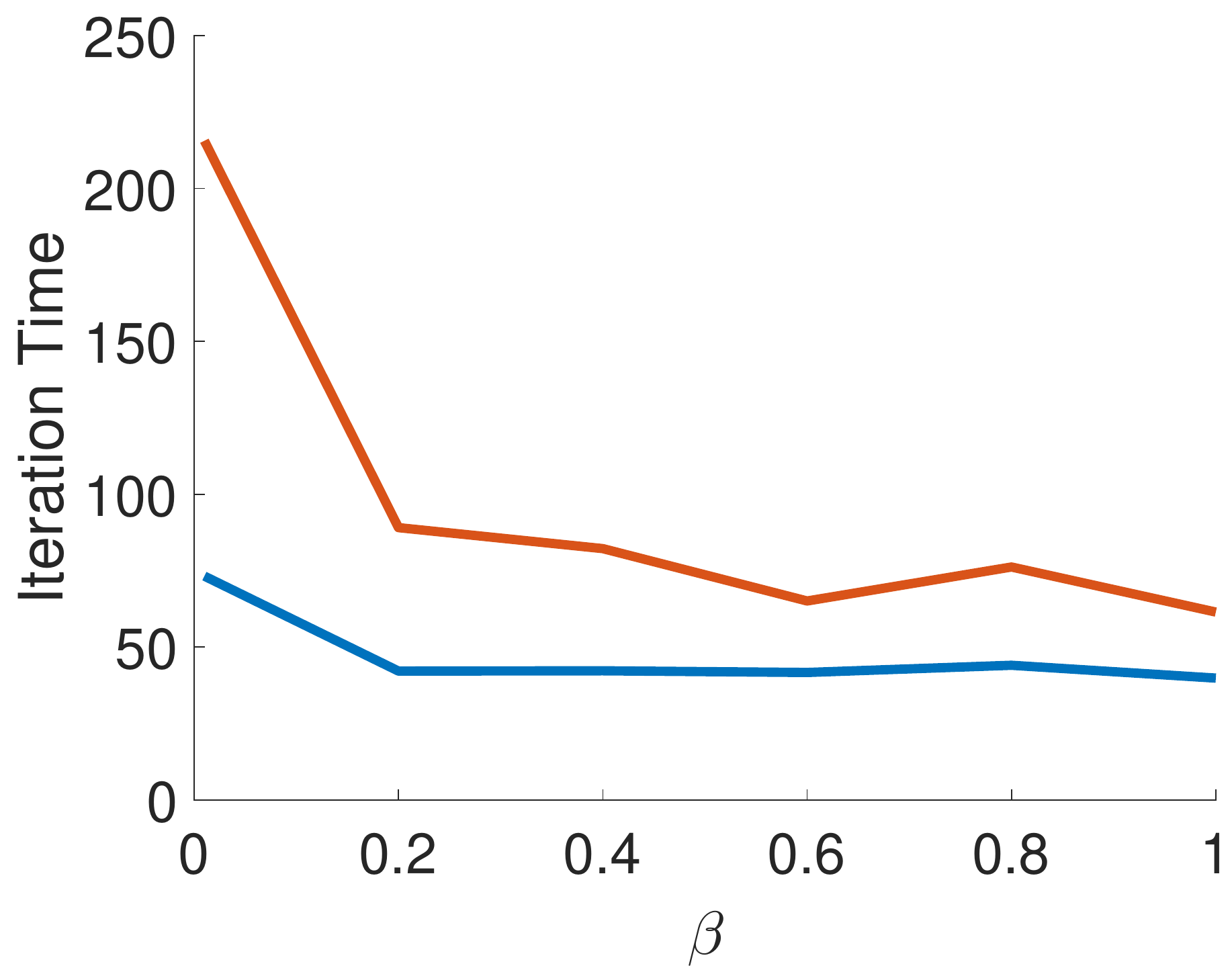}
    }
    \subfloat[][]{
        \includegraphics[width=0.315\linewidth]{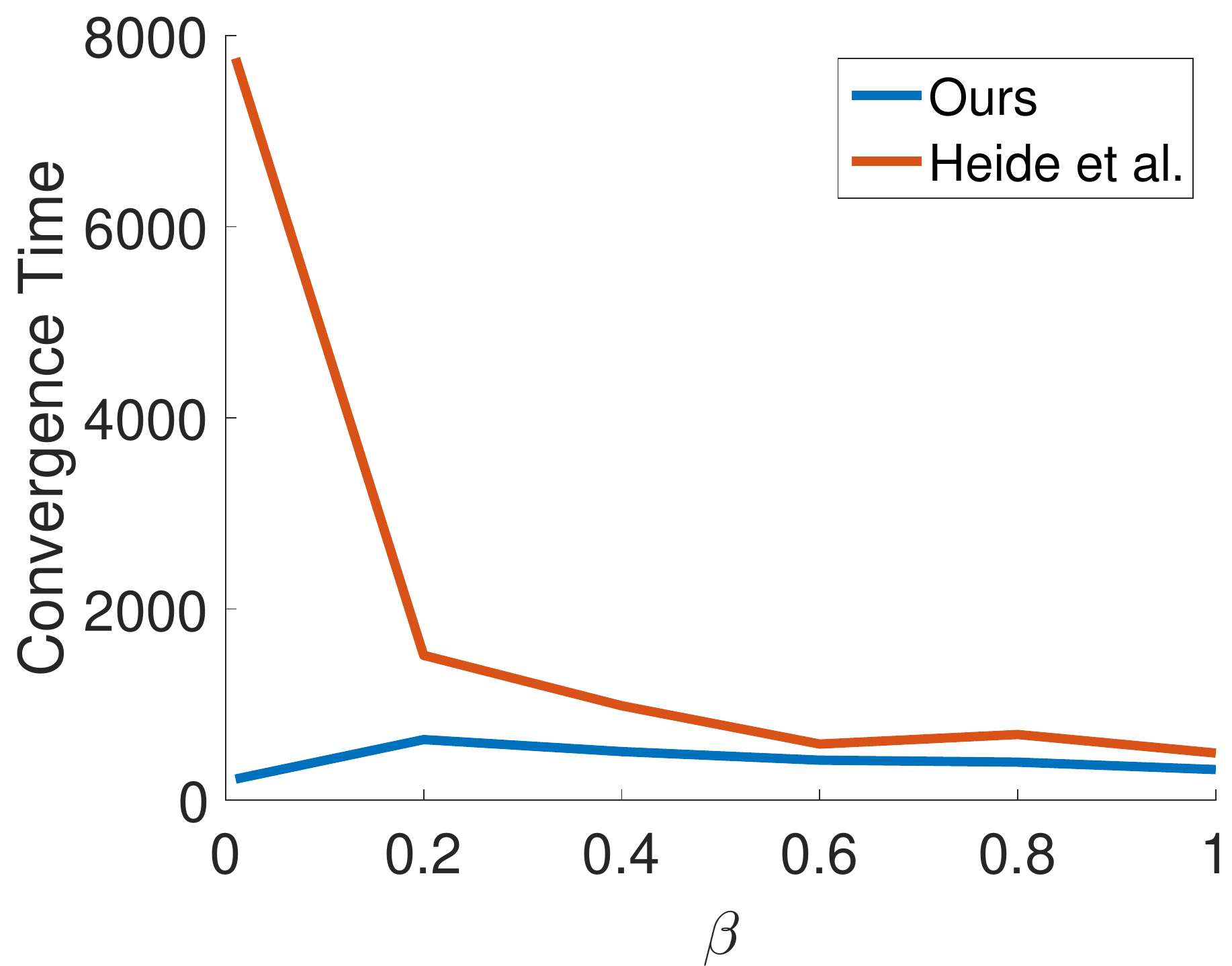}
        \label{fig:TimeIterBeta}
    }
	\caption{(a) Objective function value as a function of time for fixed $\beta$ and $K$. (b-c) Convergence time as $\beta$ is varied.}
	
\end{figure}

\begin{figure}[h]
	\centering
		\includegraphics[width=1\linewidth]{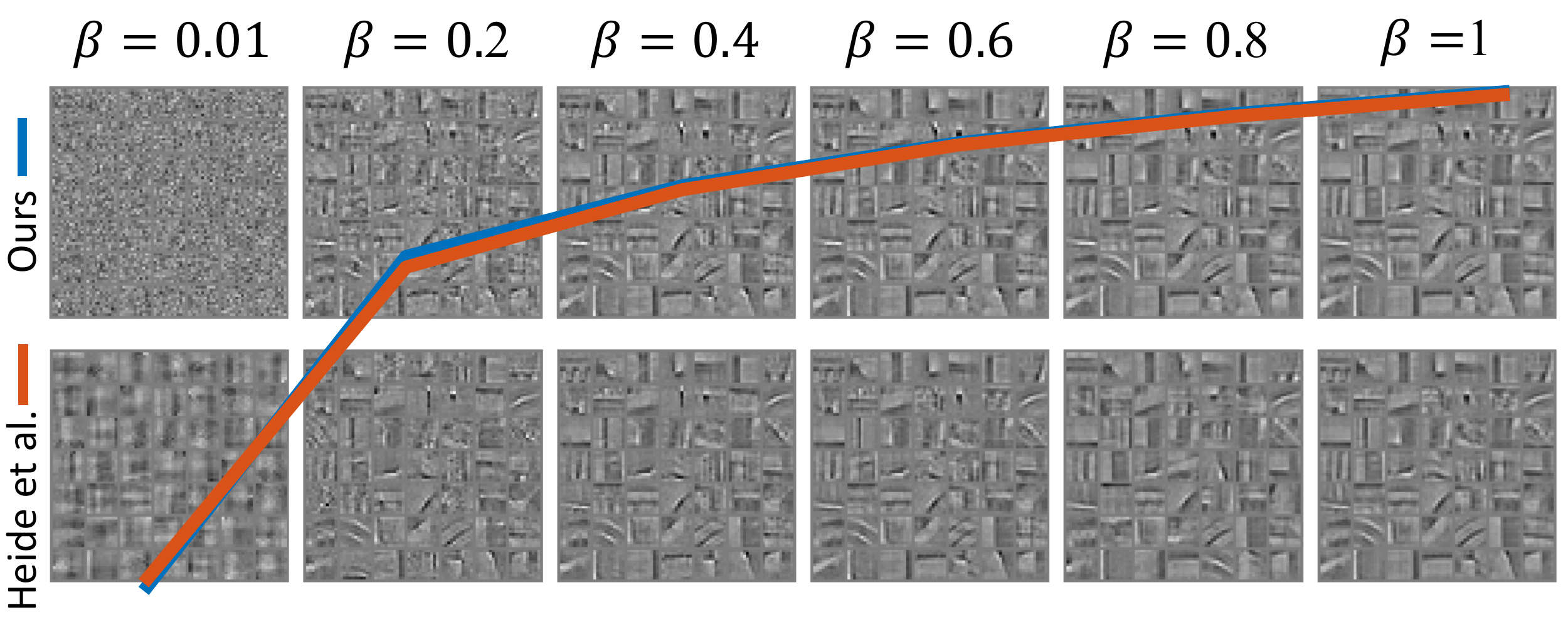}
	\centering
	\caption{Dictionary element progression as a function of $\beta$ for our method (top) and Heide \etal~\cite{Heide2015} (bottom). The curves also show the increasing objective value as $\beta$ increases.}
	\label{fig:Obj_Beta}
\end{figure}

\noindent\textbf{Learning Subproblem.} Here, our approach solves the problem iteratively using conjugate gradient to solve the linear system. Thus, the computational cost lies in solving elementwise products and divisions, with the additional cost of applying the Fourier transforms to the variables. On the other hand, Heide \etal~\cite{Heide2015} and Bibi \etal~\cite{bibi2017high} need to solve the subproblem by applying ADMM.

We observe that the performance of our dictionary learning approach improves by increasing  the number of conjugate gradient iterations involved in solving the linear system. This number decreases after each inner iteration and its cost becomes negligible within 4-6 iterations due to the warm-start initialization of $\boldsymbol{\gamma}$ at each step as shown in Figure~\ref{fig:learning_coding_conv}. On the other hand, Heide \etal~\cite{Heide2015} and Bibi \etal~\cite{bibi2017high} incur  the additional cost of solving the linear systems. Thus, our method has better scalability compared to the primal methods in which the linear systems solving step dominates as the number of images and filters increase (see section~\ref{sec:scalability}).

\begin{table}[t]
\centering
\caption{Computational complexity of the coding and learning subproblems in the primal and dual domain}
\label{complexitytable}
\begin{tabular}{c|c|c|}
\cline{2-3}
                               & \textbf{Coding Subproblem} & \textbf{Learning Subproblem} \\ \hline
\multicolumn{1}{|l|}{\textbf{Dual CSC}} & $J^2KD +JKDlogD +JKD$                 & $ Q(JKD +JKDlogD)$                   \\ \hline
\multicolumn{1}{|l|}{\textbf{CSC}~\cite{Heide2015}}      & $JKD +JKDlogD +JKD$                 & $JK^2D+ JKDlogD+JKD$                   \\ \hline
\multicolumn{1}{|l|}{\textbf{TCSC}~\cite{bibi2017high}}     & $K^3D +JKDlogD +JKD $               & $ K^3D+JKDlogD+JKD$                   \\ \hline
\end{tabular}
\end{table}

\subsection{CSC Convergence}
\label{sec:convergence}

\begin{figure*}[t]
	\subfloat[][]{
        \includegraphics[width=0.315\linewidth]{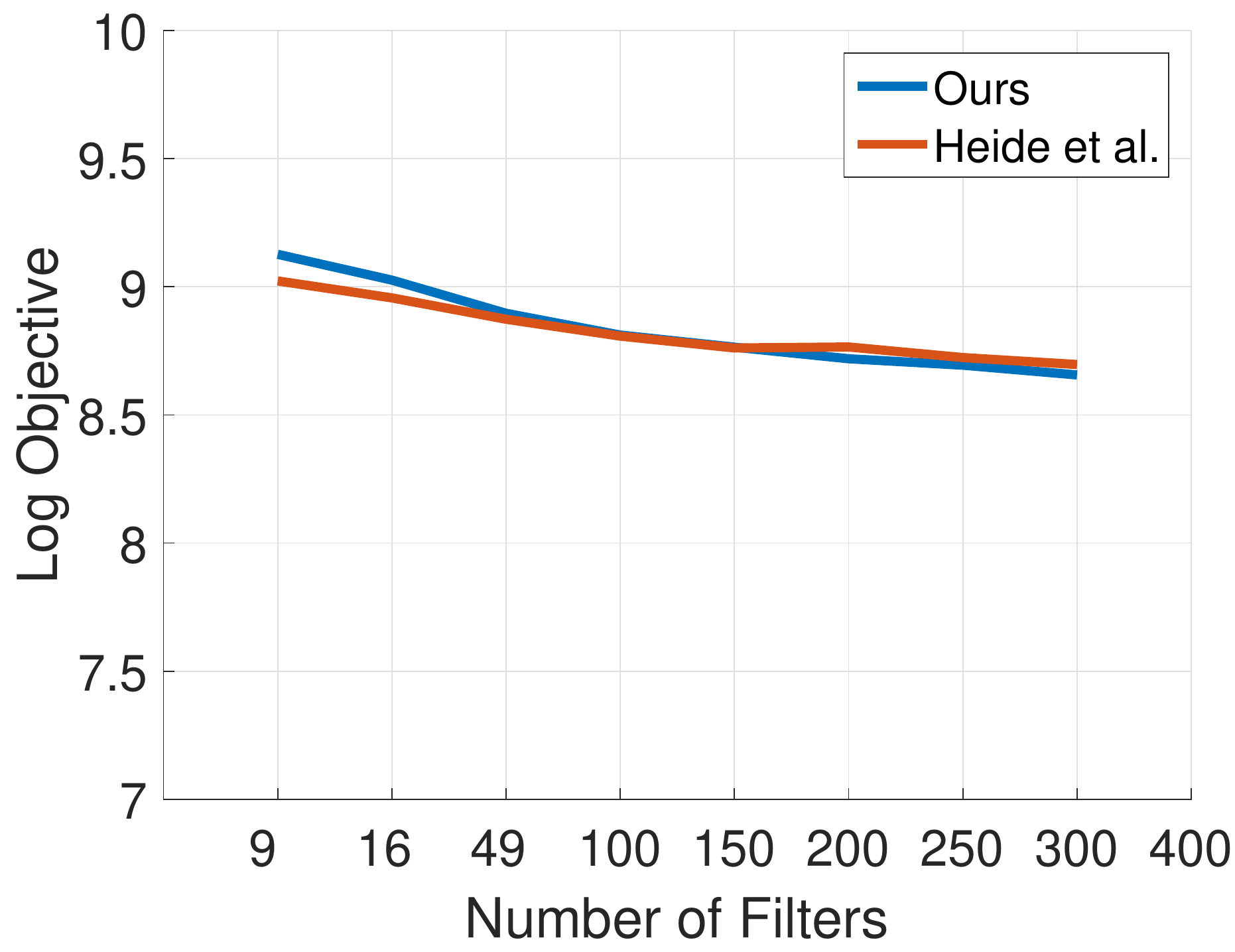}
    }
    \subfloat[][]{
        \includegraphics[width=0.315\linewidth]{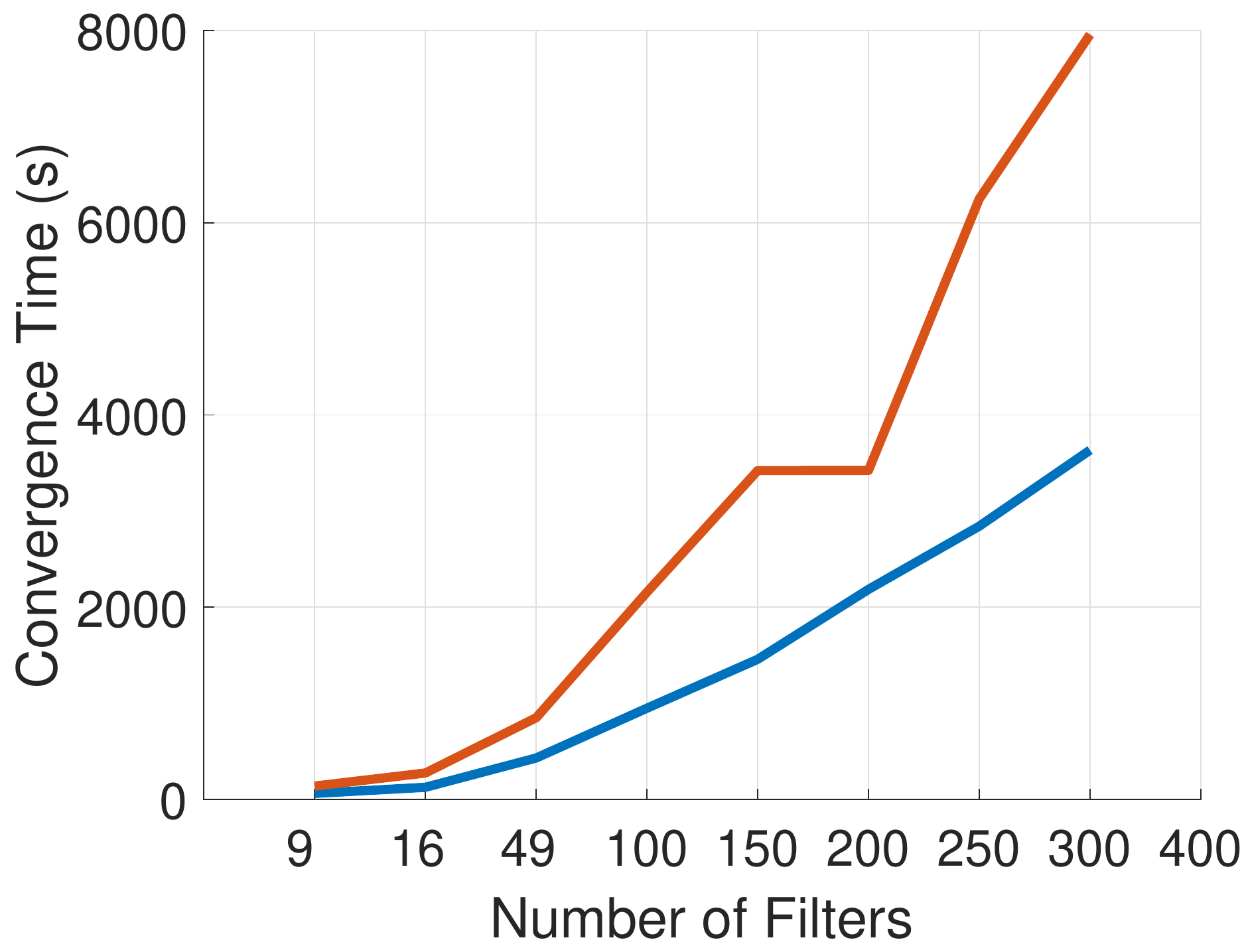}
    }
    \subfloat[][]{
        \includegraphics[width=0.315\linewidth]{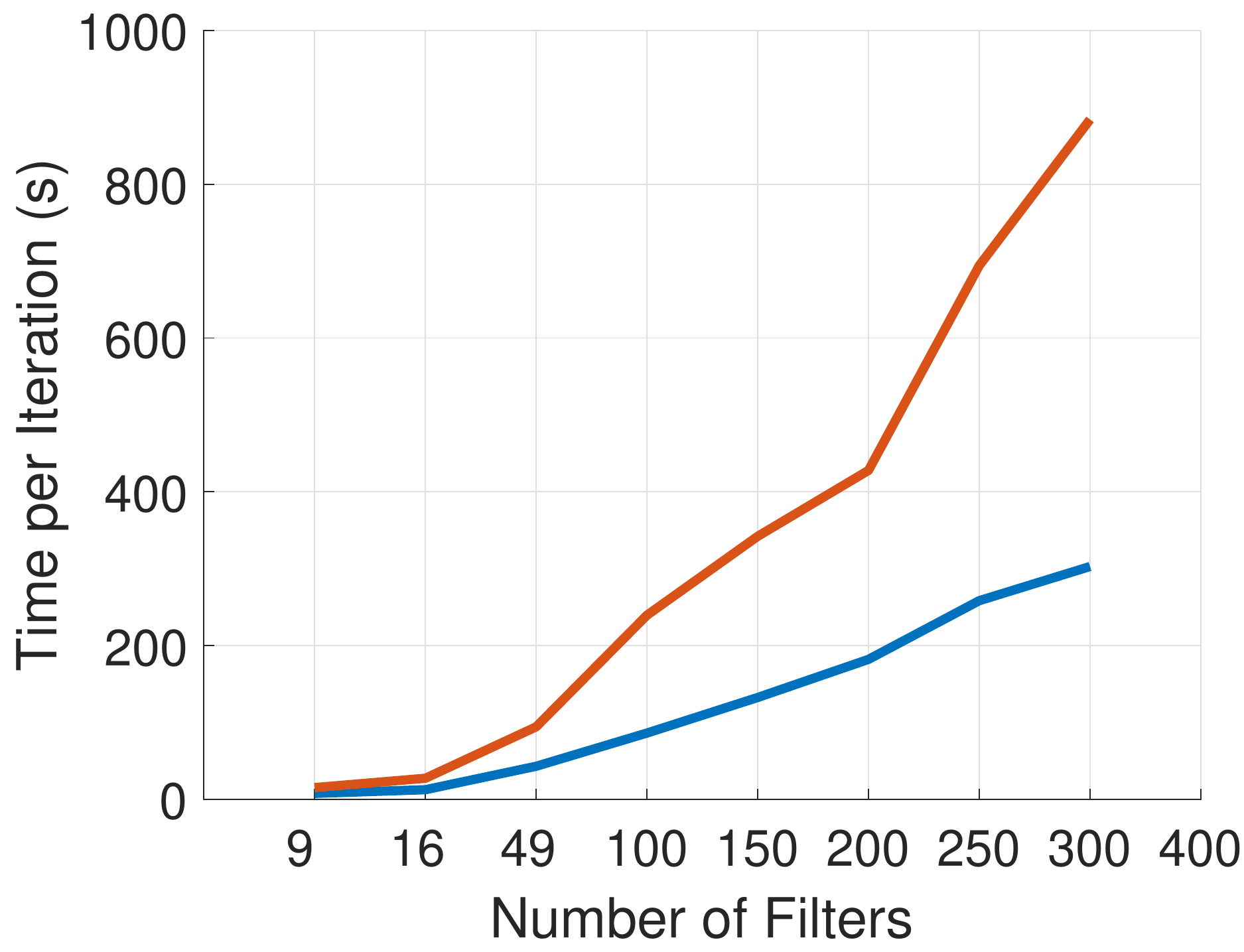}
    }

    \subfloat[][]{
        \includegraphics[width=0.315\linewidth]{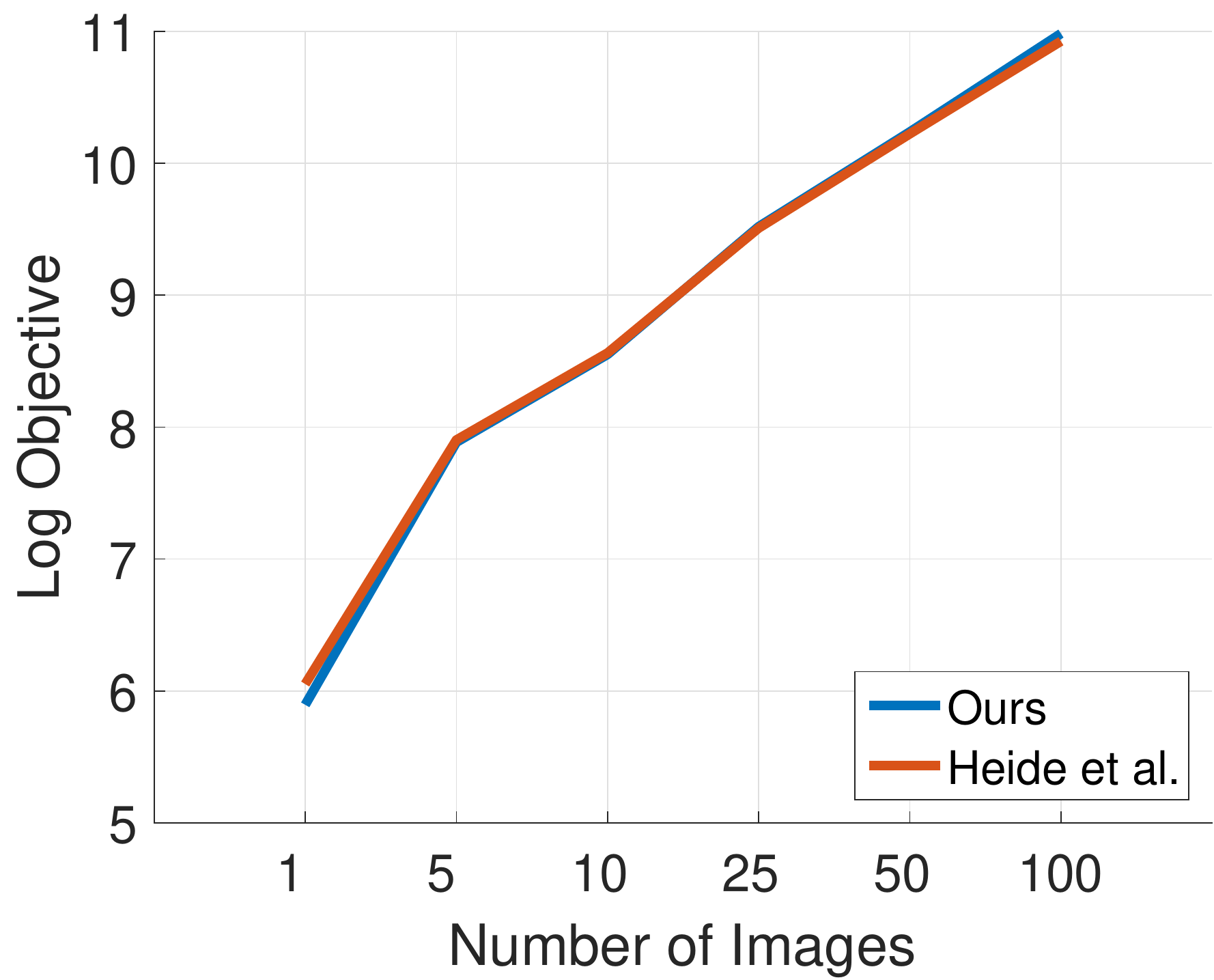}
    }
    \subfloat[][]{
        \includegraphics[width=0.315\linewidth]{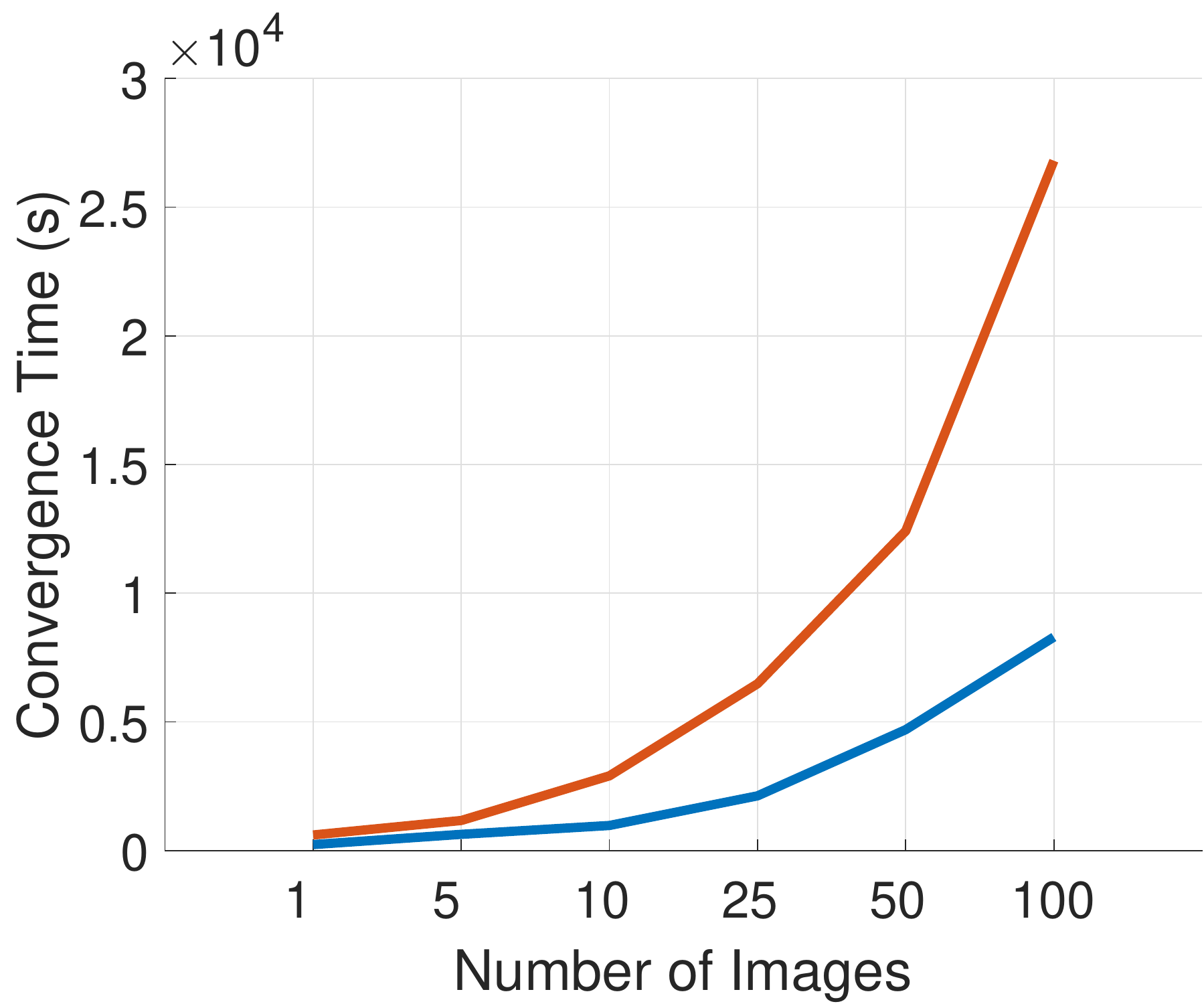}
    }
    \subfloat[][]{
        \includegraphics[width=0.315\linewidth]{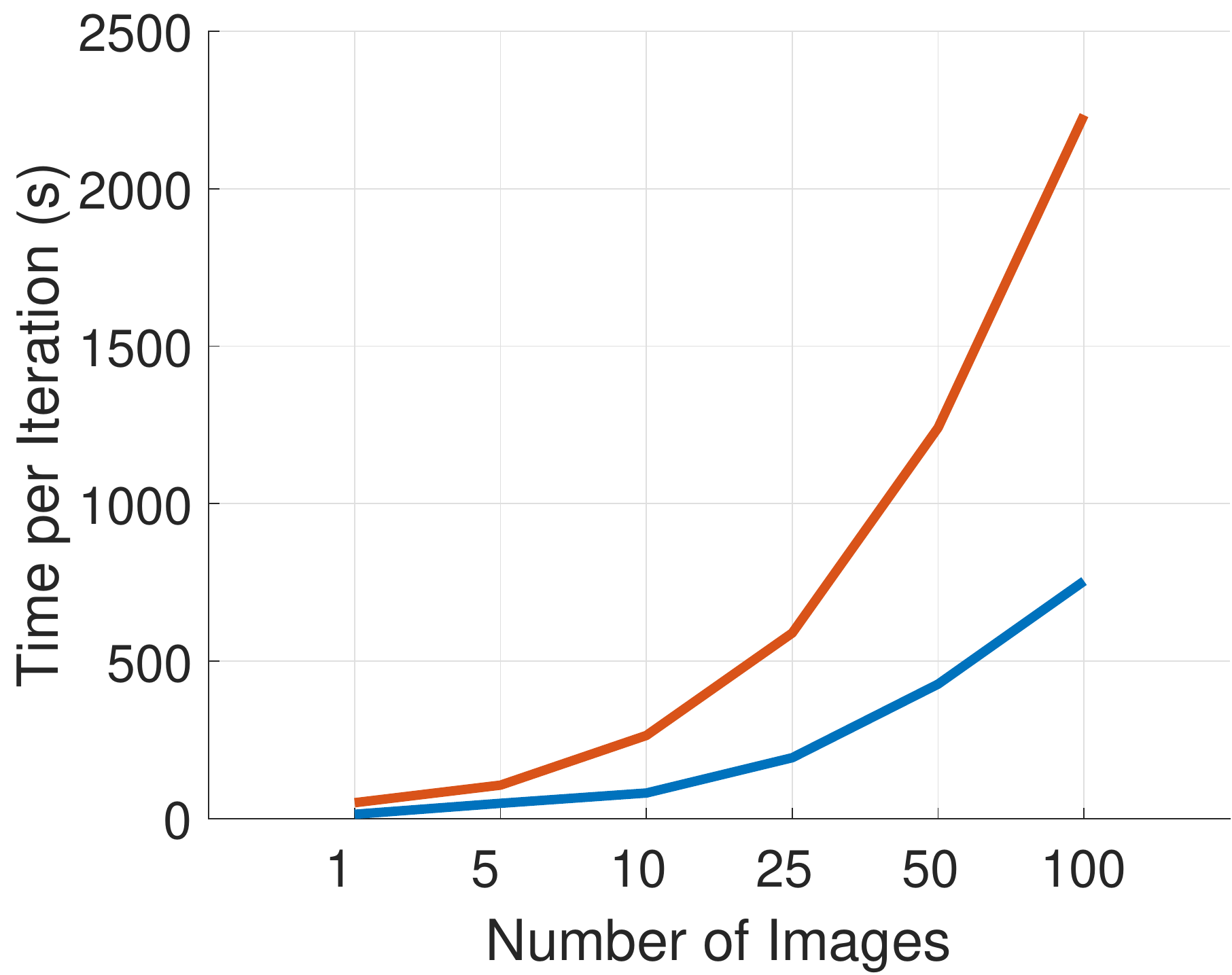}
    }
	\caption{Objective function value (a,d), convergence time (b,e), and average iteration time in seconds (c,f) as a function of number of dictionary elements (top) and number of images (bottom).}
	\label{fig:TimeIterK}
\end{figure*}

In this section, we analyse the convergence properties of our approach to regular convolutional sparse coding. In Figure~\ref{fig:Obj_Iter}, we plot the convergence of our method compared to the state of the art~\cite{Heide2015} on the \emph{city} dataset for fixed $\beta~=~0.5$ and $K~=~49$. We also show the progression of the learnt filters in correspondence with the curves. As shown in the figure, the two methods converge to the same solution. Figure~\ref{fig:Obj_Time} plots how the objective value decreases with time for each of the two methods using the same parameters as above. This shows that our method converges significantly faster than~\cite{Heide2015}.

\begin{figure}[t]
	\centering
	\subfloat[][]{
        \includegraphics[width=0.315\linewidth]{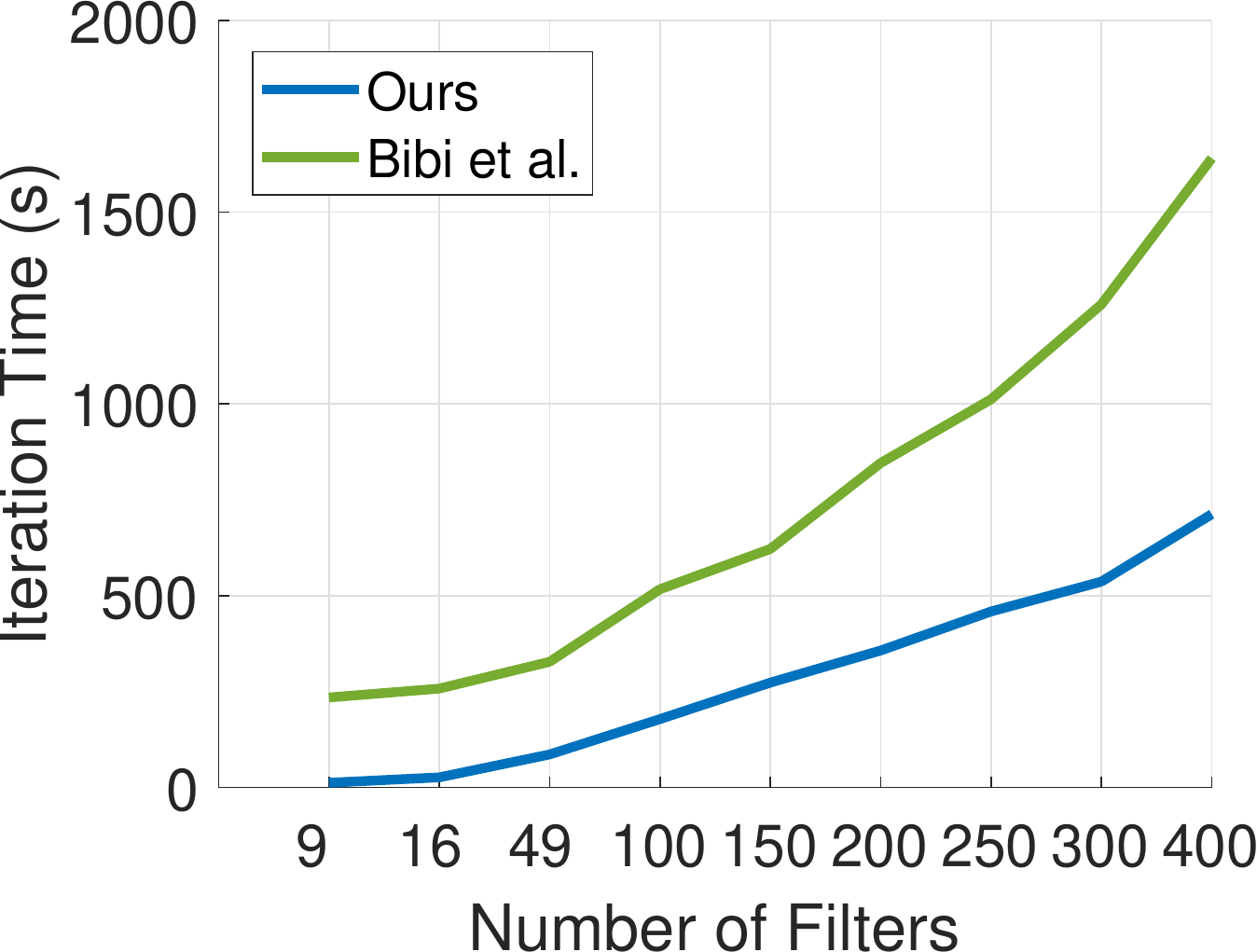}
        \label{fig:3D_K_Image}
    }
	\subfloat[][]{
        \includegraphics[width=0.315\linewidth]{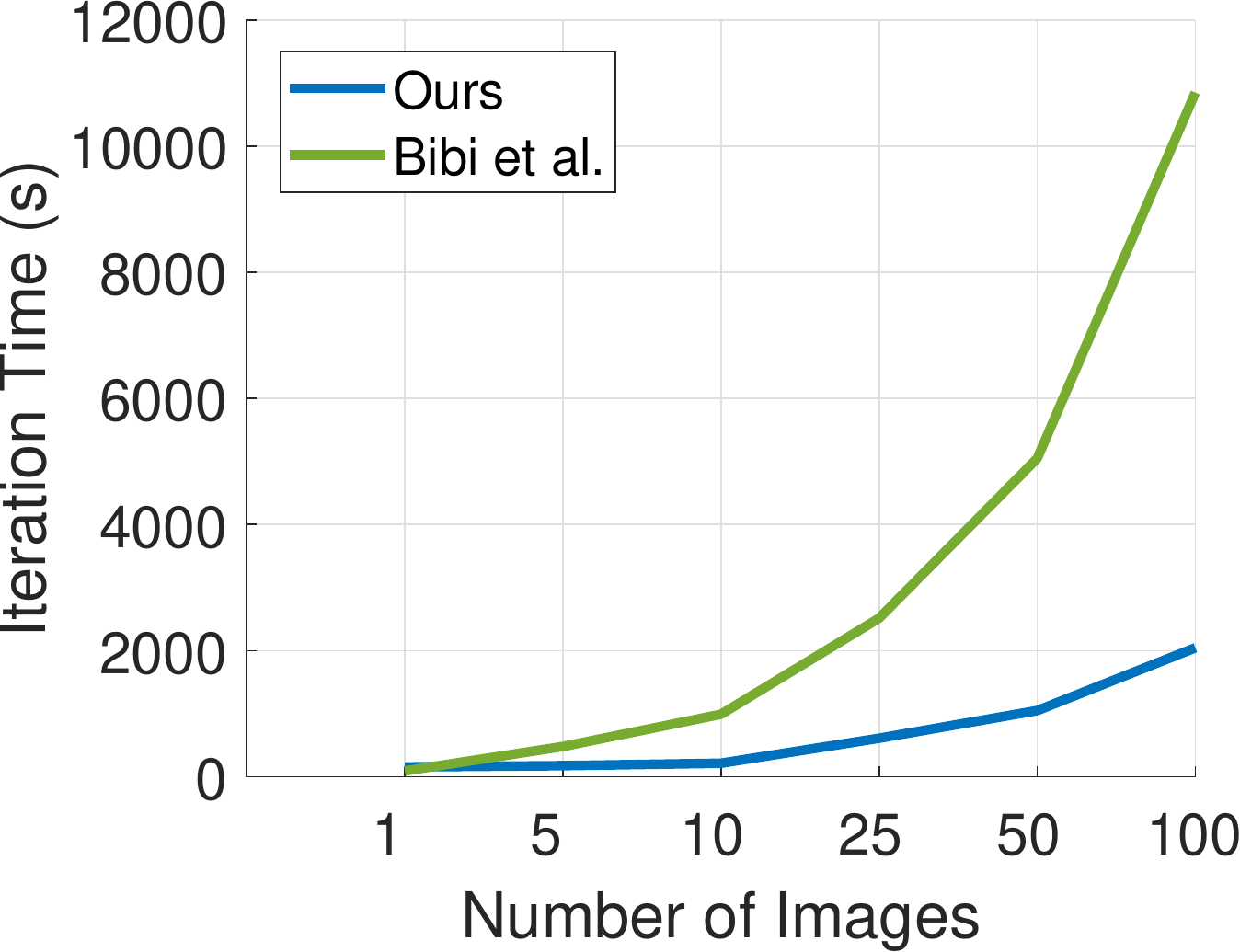}
        \label{fig:3D_N_Image}
    }
    \subfloat[][]{
        \includegraphics[width=0.315\linewidth]{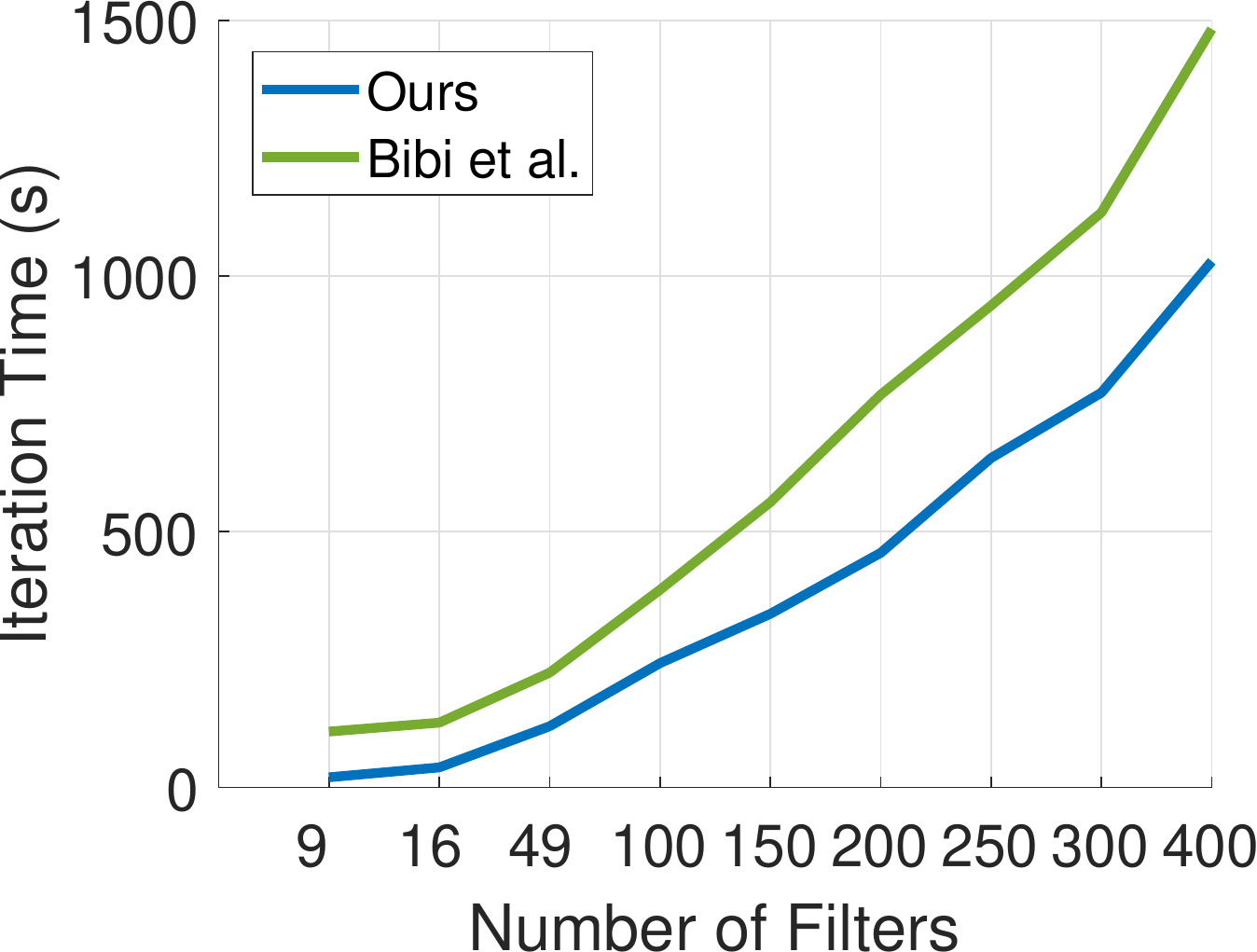}
        \label{fig:4D_K_Video}
    }
	\caption{(a) Convergence time on colored images as $K$ is varied. (b) Convergence time on colored images as $N$ is varied. (c) Convergence time on colored video as $K$ is varied. }
	\label{fig:tcsc}
\end{figure}

\begin{figure}[t]
	\centering
	\subfloat[][]{
        \includegraphics[width=0.315\linewidth]{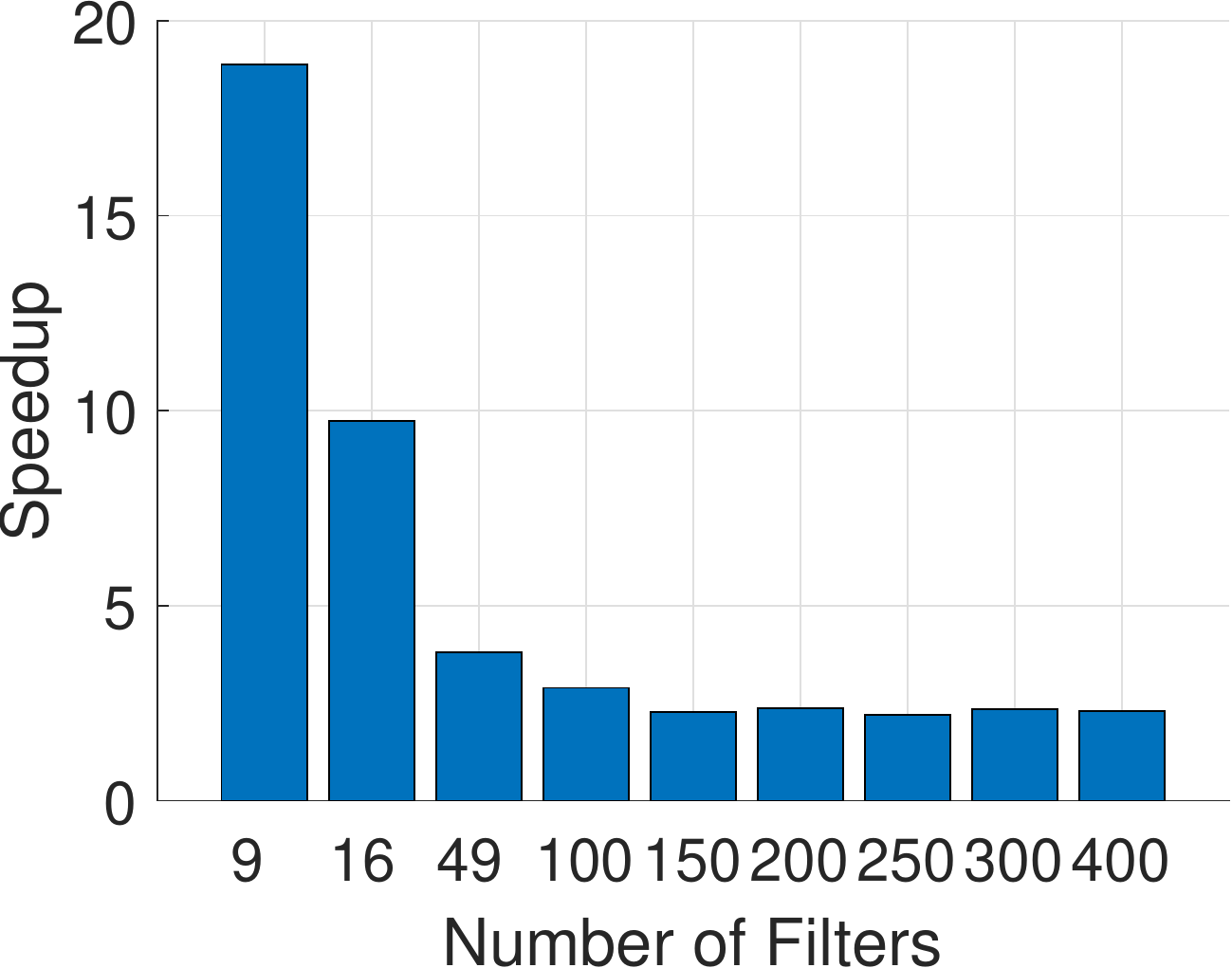}
        \label{fig:3D_K_speedup}
    }
	\subfloat[][]{
        \includegraphics[width=0.315\linewidth]{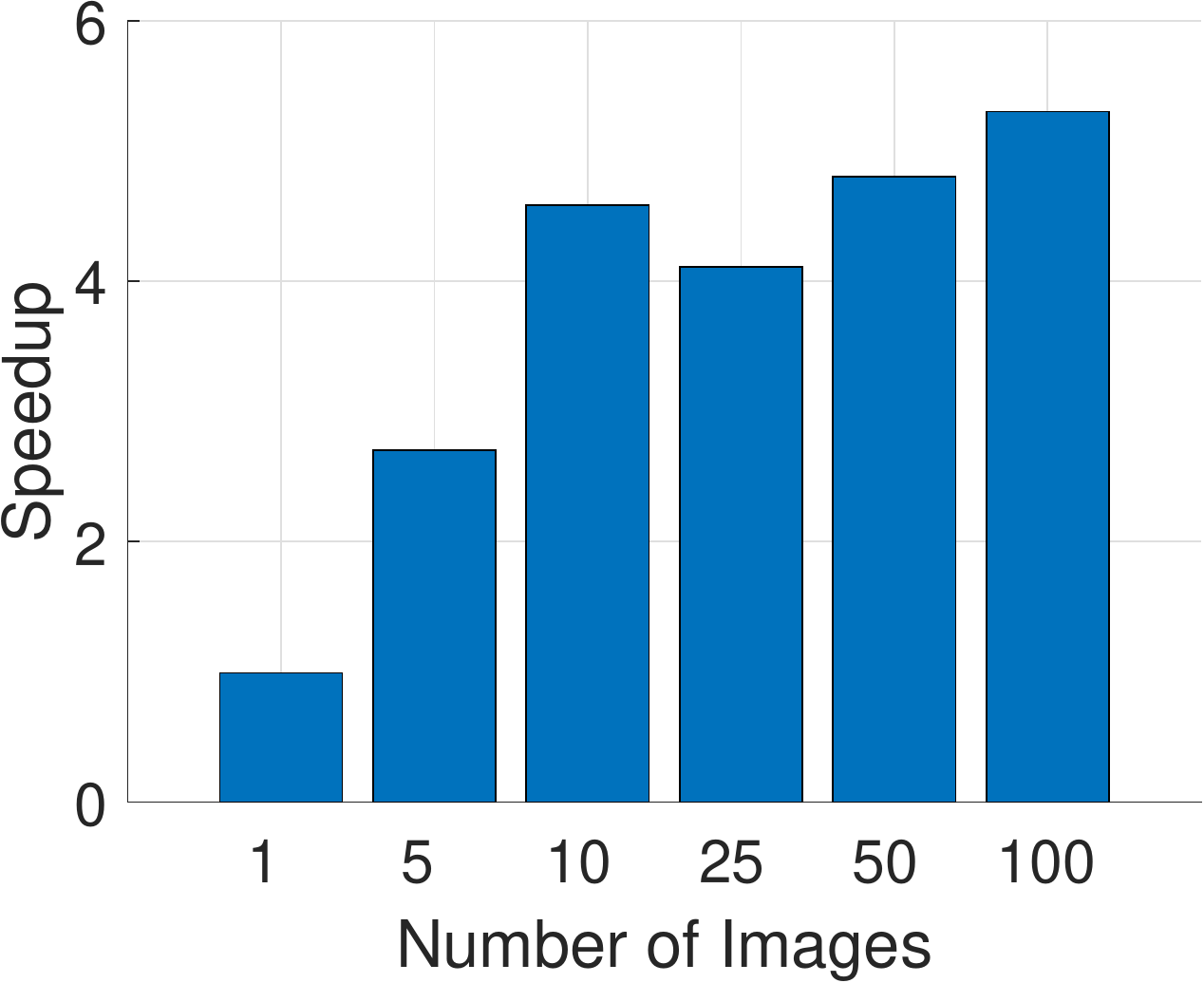}
        \label{fig:3D_N_spedup}
    }
    \subfloat[][]{
        \includegraphics[width=0.315\linewidth]{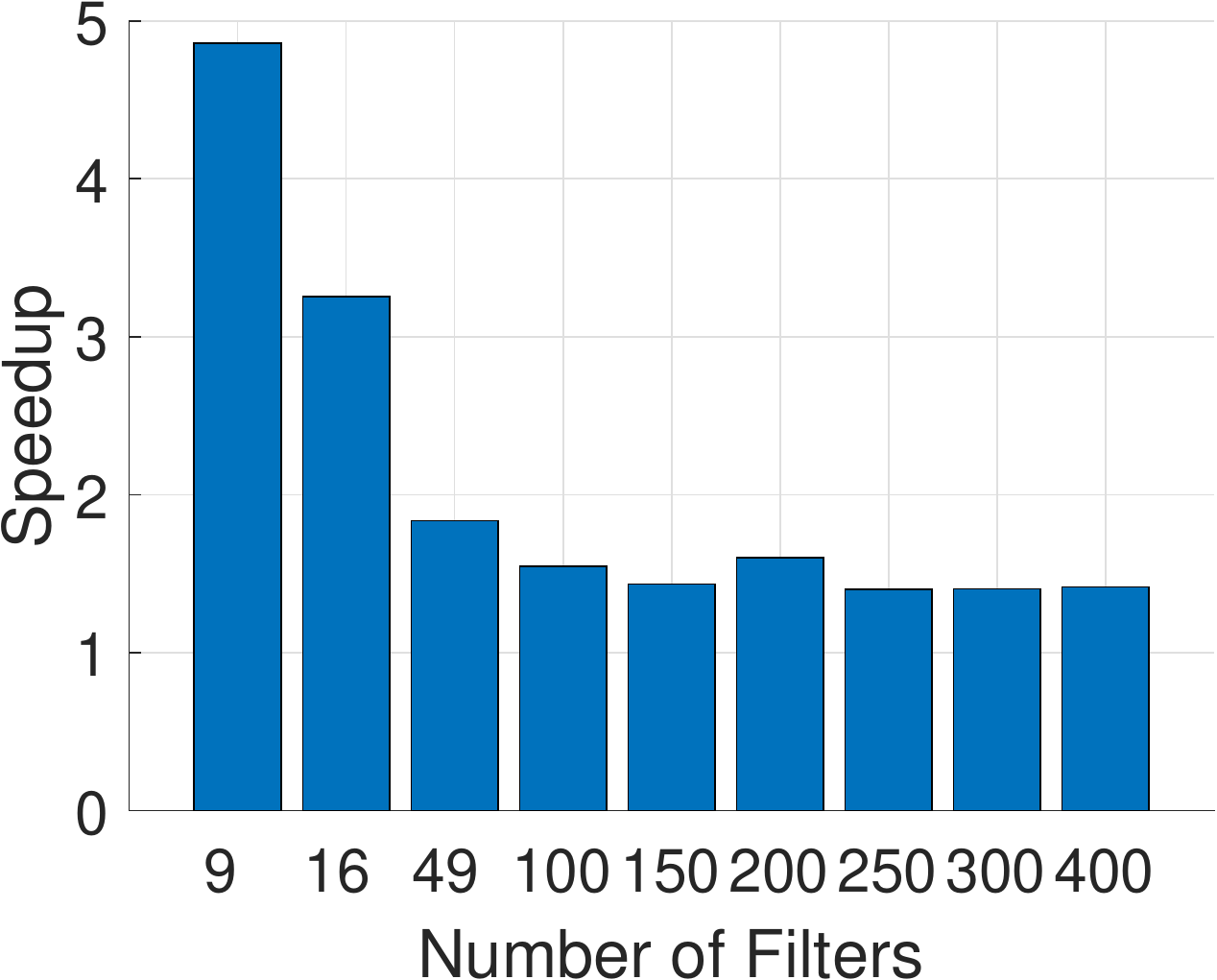}
        \label{fig:4D_K_speedup}
    }
	\caption{(a) Speedup on colored images as $K$ is varied. (b) Speedup on colored images as $N$ is varied. (c) Speedup on colored video as $K$ is varied. }
	\label{fig:tcsc_speedup}
\end{figure}


We also plot in Figure~\ref{fig:Obj_Beta} the objective value as a function of the sparsity coefficient $\beta$. The plot shows how increasing the sparsity coefficient results in an increase in the objective value, and more importantly, it verifies that our method converges to an objective value similar to that of~\cite{Heide2015} even though we reach a solution faster as shown in Figure~\ref{fig:TimeIterBeta}. Figure~\ref{fig:Obj_Beta} also shows how the dictionary elements vary with  $\beta$.

\subsection{Scalability}
\label{sec:scalability}

In this section, we analyze the scalability of the CSC problem with increasing number of filters and images. We compare our approach to Heide et. al.~\cite{Heide2015} in terms of the overall convergence time and average time per iteration for reaching the same final objective value. To ensure that the two problems achieve similar overall objective values, we make sure that each of the methods runs until convergence for both the coding and learning convex subproblems with the same initial point. Figure~\ref{fig:TimeIterK} shows that the computation time of each of the methods increases when the number of filters and images increases. It also shows a significant speedup (about 2.5x) for our approach over that of~\cite{Heide2015}.

\subsection{TCSC Results}


In this section, we show results for TCSC on color images and videos. We compare the performance of our dual formulation for TCSC with that of Bibi \etal~\cite{bibi2017high}. Figures~\ref{fig:3D_K_Image} and \ref{fig:3D_N_Image} show iteration time results on color images from the \emph{fruit} and \emph{house} datasets respectively with varying the number of filters and number of images. Correspondingly, we show the speedup acheived by our dual formulation for images and videos in Figure~\ref{fig:tcsc_speedup}. As shown, our dual formulation achieves up to 20 times speed-up compared to the primal solution. We can observe higher speedups for smaller number of filters and we can also observe that the speedup is approximately constant as the number of images increases. This is inline with our complexity analysis in which we verified that Sherman Morrison formula is no longer applicable in the primal domain for TCSC, while parallelization is still applicable in the dual. \\

We also show in Figure~\ref{fig:4D_K_Video} the iteration time as the number of filters varies on colored video. Since the video is considered as a 5D tensor, 3D Fourier transforms are applied resulting in a lower speedup but still maintaining a lower computation time compared to~\cite{bibi2017high}.


\section{Conclusion and Future Work}
We proposed our approach for solving the convolutional sparse coding problem by posing and solving each of its underlying convex subproblems in the dual domain. This results in a lower computational complexity than previous work. We can also easily extend our proposed solver to CSC problems of  higher dimensional data. We demonstrated that tackling CSC in the dual domain results in up to 20 times speedup compared to the current state of the art.
In future work, we would like to experiment with additional regularizers for CSC. We could make use of the structure of the input signal and map the regularizer over to the sparse maps to reflect this structure. For example, for images with a repetitive pattern, a nuclear norm can be added as a regularizer, which is equivalent to making the sparse maps low rank.

\bibliographystyle{splncs}
\bibliography{./source/egbib}
\end{document}